\documentclass[lettersize,journal]{IEEEtran}
\usepackage{caption}
\usepackage{amsmath,amsfonts}
\usepackage{algorithm}
\usepackage{tabularray}%20240123
\usepackage{color}%20240123
\usepackage{array}
\usepackage{textcomp}
\usepackage{stfloats}
\usepackage{float}
\usepackage{graphicx}
\usepackage{cite}
\usepackage{subfig}
\usepackage{caption}
\usepackage{subcaption}
\usepackage{makecell}%20230527
\usepackage{booktabs}%20230523 
\usepackage{multirow}%20230523
\usepackage{algorithmicx}
\usepackage[noend]{algpseudocode}
\usepackage[table,xcdraw]{xcolor}
\usepackage[table,xcdraw]{xcolor}
\usepackage{color}
\usepackage{lineno}
\usepackage{nomencl}

\usepackage{hyperref} %网页超链接

\usepackage{arydshln}%虚线

\usepackage{amsthm}   % 这个包用于定理环境
\theoremstyle{plain}

\usepackage{cleveref}
\crefname{table}{Table}{Tables}
\crefname{assumption}{Assumption}{Assumptions}
\crefname{section}{Section}{Sections}
\crefname{appendix}{Appendix}{Appendixes}
\crefname{algorithm}{Algorithm}{Algorithms}
\crefname{figure}{Figure}{Figures}
\crefname{theorem}{Theorem}{Theorems}
\crefname{lemma}{Lemma}{Lemmas}
\crefname{definition}{Definition}{Definitions}
\renewcommand{\eqref}[1]{Eq.~(\textup{\ref{#1}})}

\usepackage{bm} % 包含 bm 包以支持加粗

\usepackage{algorithm}
\usepackage{algorithmicx}
\usepackage{algpseudocode}
\usepackage{amsthm}   % 这个包用于定理环境
\theoremstyle{plain}

\usepackage{xcolor}
\usepackage{todonotes}

\usepackage{cleveref}
\crefname{table}{Table}{Tables}
\crefname{assumption}{Assumption}{Assumptions}
\crefname{section}{Section}{Sections}
\crefname{appendix}{Appendix}{Appendixes}
\crefname{algorithm}{Algorithm}{Algorithms}
\crefname{figure}{Figure}{Figures}
\crefname{theorem}{Theorem}{Theorems}
\crefname{lemma}{Lemma}{Lemmas}
\crefname{definition}{Definition}{Definitions}
\renewcommand{\eqref}[1]{Eq.~(\textup{\ref{#1}})}
\usepackage{bm} % 包含 bm 包以支持加粗
\makenomenclature

\begin{document}
%\pagewiselinenumbers
%\switchlinenumbers
\graphicspath{ {./picture/} }
\bibliographystyle{IEEEtran}

\title{YOLO-MST: Multiscale deep learning method for infrared small target detection based on super-resolution and YOLO}
\author{ Taoran Yue$^{1}$, Xiaojin Lu$^{2}$, Jiaxi Cai$^{2}$, Yuanping Chen$^{1*\ddagger}$, Shibing Chu$^{1*\ddagger}$ 
\thanks{$^{\ddagger}$Corresponding author.}      
\thanks{This work gratefully acknowledges the National Natural Science Foundation of China (No. 11904137, 12074150 and 12174157) and the financial support from Jiangsu University (No. 4111190003). ({\emph{ Corresponding author: Yuanping Chen Shibing Chu.}})
}
\thanks{$^{1}$Jiangsu Engineering Research Center on Quantum Perception and Intelligent Detection of Agricultural Information, Zhenjiang, 212013, China.(e-mail:chenyp@ujs.edu.cn)}
\thanks{$^{2}$School of Physics and Electronic Engineering, Jiangsu University, Zhenjiang, 212013, China.(e-mail:chenyp@ujs.edu.cn)}}

\maketitle

\begin{abstract} With the advancement of aerospace technology and the increasing demands of military applications, the development of low false-alarm and high-precision infrared small target detection algorithms has emerged as a key focus of research globally. However, the traditional model-driven method is not robust enough when dealing with features such as noise, target size, and contrast. The existing deep-learning methods have limited ability to extract and fuse key features, and it is difficult to achieve high-precision detection in complex backgrounds and when target features are not obvious. To solve these problems, this paper proposes a deep-learning infrared small target detection method that combines image super-resolution technology with multi-scale observation. First, the input infrared images are preprocessed with super-resolution and multiple data enhancements are performed. Secondly, based on the YOLOv5 model, we	proposed a new deep-learning network named YOLO-MST. This network includes replacing the SPPF module with the self-designed MSFA module in the backbone, optimizing the neck, and finally adding a multi-scale dynamic detection head to the prediction head. By dynamically fusing features from different scales, the detection head can better adapt to complex scenes. The mAP@0.5 detection rates of this method on two public datasets, SIRST and IRIS, reached 96.4\% and 99.5\% respectively, more effectively solving the problems of missed detection, false alarms, and low precision.

\end{abstract}

\begin{IEEEkeywords}
Deep learning, Infrared remote sensing, Small target recognition, Super-resolution, YOLO
\end{IEEEkeywords}

\maketitle

\section{Introduction}
Infrared remote sensing technology has advanced significantly, finding applications in military, civilian, and public security fields. In the military, it supports tasks like nighttime ship inspections \cite{shao2021vessel} and early warning systems \cite{bertrand2020infrared}. Civilian uses include disaster early warning \cite{zhang2010applications}, crop monitoring \cite{gerhards2019challenges}, and equipment fault detection \cite{zheng2022lightweight}. In public safety, it enhances disaster rescue efforts \cite{yadav2022thermal} by efficiently locating trapped individuals and reducing casualties. Besides, deep-learning-based target detection technology has found broad applications in various fields, including underwater image analysis \cite{wang2024deep, li2021lidar}, radar detection, anti-interference systems, and polarization imaging. Unlike visible light, infrared radiation excels in low-light and adverse weather conditions due to its high penetration and temperature sensitivity \cite{sousa2020thermal}, making it indispensable for small target detection across various scenarios.
\newline \indent Infrared small targets, characterized by low contrast, weak signals, and small size \cite{chapple1999target}, pose significant detection challenges due to their interaction with complex backgrounds (e.g., clouds, waves, buildings) and high noise levels. Weak signals are often masked by background noise, reducing detection sensitivity. Improving the algorithm's ability to differentiate targets from the background, enhance noise robustness, and increase detection accuracy remains a critical focus in infrared small target detection research.
\newline \indent To solve these problems, this paper proposes an infrared small target detection (ISOD) method based on YOLOv5 called YOLO-MST. Specifically, before inputting the infrared images into the detection network, a super-resolution model named ESRGAN  \cite{wang2021real}  is employed to preprocess the images, enhancing the  network's detection accuracy. Additionally, to address the feature loss caused by over-sampling of small targets \cite{wang2021multi}, we have designed a multi-Scale Feature Aggregation (MSFA) module to replace the Spatial Pyramid Pooling (SPPF) module in the backbone, enabling richer feature extraction through multi-scale observation and thereby reducing false alarms. In addition, we added a convolution operations to the neck and deleted the large target output section, so that the model can focus more on small target detection and reduce interference. Then, in the prediction head, we incorporated the DyHead detection head \cite{dai2021dynamic},which addresses the challenges of small target detection in infrared images through dynamic feature fusion. This approach is particularly effective in scenarios involving complex backgrounds or severe interference, significantly enhancing both the accuracy and stability of the model. Finally, we verified the effectiveness of the proposed YOLO-MST method by comparing it with various SOTA methods. 
\newline \indent The contributions of this paper include:
\newline \indent(1) Before inputting the infrared images into the proposed detection network YOLO-MST, the ESRGAN \cite{wang2021real} (super-resolution reconstruction) is employed to preprocess the input images, effectively enhancing the resolution and detail of target features in infrared images, thus significantly improving the model's detection accuracy.
\newline \indent(2) In the backbone, we designed a MSFA module to replace the SPPF module, capturing the multi-scale feature information of the infrared images through three atrous convolutions with different rates, and finally performing weighted fusion to understand the multi-scale feature of the images. 
\newline \indent(3) In the neck, we deleted the output part of large target detection and added a layer of Conv convolution between the two outputs to optimize the neck structure of the model, so that the model can focus more on small target detection.
\newline \indent(4) In the prediction head, we added the DyHead \cite{dai2021dynamic} detection head. It contains three modules: scale attention, spatial attention, and task attention. It fuses features of different scales according to semantic importance, then uses deformable convolution to focus on the shape of the object and enhance feature extraction. Finally, the dynamic switching of feature channels is implemented to adapt to the requirements of different tasks.
\newline \indent Experimental results show that the mAP@0.5 detection rates on two public datasets SIRST and IRIS reach 96.4\% and 99.5\% respectively. Compared with the existing SOTA target detection methods, YOLO-MST performs well.
\newline \indent The remainder of this paper is organized as follows: Section 2 reviews related work, Section 3 presents our dataset preprocessing and detection network, Section 4 discusses experimental details, results, and analysis, and Section 5 concludes the paper.

\section{Related work} %%2%%

\subsection{Traditional infrared small target detection method}    %%2.1%%
%%\label{Traditional infrared small target detection method}
Traditional small target detection methods include filtering, human visual system (HVS), and low-rank representation. Among them, the filtering-based methods include top-hat filtering \cite{bai2010analysis}, maximum median and maximum mean filtering \cite{deshpande1999max}, spatial domain high-pass filtering method \cite{jia1999infrared}, bilateral filtering method \cite{bae2010small}, two-dimensional minimum mean square filtering method \cite{bae2012edge}, wavelet transform method \cite{sun2010dim}, etc. However, these methods are limited by smooth and slowly changing backgrounds. They are not robust enough to changes in target size and can only suppress uniform background clutter but not complex background noise.

\subsection{Methods based on deep-learning}   %%2.2%%
%%\label{Deep learning based methods}
According to different processing paradigms, they can be divided into detection-based methods and segmentation-based methods. Detection-based methods are mainly divided into two-stage algorithms and one-stage algorithms. Two-stage algorithms include R-CNN \cite{lee2017me}, FAST R-CNN \cite{girshick2015fast} and Faster R-CNN \cite{ren2015faster}. One-stage algorithms include single-shot multi-frame detectors \cite{liu2016ssd}, RetinaNet \cite{lin2017focal} and YOLO series. Liu et al. \cite{liu2017image} first introduced CNN and used CNN to generate infrared target samples with controllable signal-to-noise ratios, but the targets they detected were not real targets in the natural environment. Before the emergence of YOLO, the mainstream target detection method was to traverse each part of the original image one by one through sliding windows of different sizes to determine whether the area detected by the classifier contains the target. Although this method has clear logic, and is extremely slow because it needs to calculate every position in the image.
\newline \indent To efficiently solve these problems, Redmond et al. \cite{redmon2016you} proposed an end-to-end target detection network called YOLO (You Only Look Once) in 2016. This method revolutionized object detection by replacing the traditional sliding window approach with a single regression process, analyzing the entire image in one pass, YOLO can significantly reduce the time required for target detection and perform well in terms of accuracy. With the introduction of YOLOv1, single-stage target detection methods have gradually gained attention. The YOLO network subsequently released several updated versions,ranging from 1.0-10.0 \cite{redmon2017yolo9000,farhadi2018yolov3,bochkovskiy2020yolov4,li2022yolov6,wang2023yolov7,wang2024yolov10}. The continuous development of these versions combines factors such as detection accuracy, detection speed, and network scale, enriching the use of the YOLO series of models. YOLOv5 is considered to be the most advanced deep-learning network for infrared small target detection in the past two years, especially on conventional datasets. Mou et al. \cite{mou2023yolo} designed YOLO-FR, a network focused on infrared small target detection. Ronghao Li et al. \cite{li2023yolosr} proposed a video target detection method based on super-resolution and deep-learning, named YOLOSR-IST, and introduced the Swin Transformer Block to replace the bottleneck layer in the network C3 module. In 2024, Hao et al. \cite{hao2024infrared} proposed the YOLO-SR model, which introduced the BTB and C3-Neck modules. Compared with the current advanced ISOD method, this method more effectively solves the problems of missed detection and false alarms. At the same time, these research results enrich the application scenarios of the YOLO series models and provide more options for weak infrared small target detection tasks.

%\balance
\begin{figure*}[ht]
	\centering
	\includegraphics[width=\textwidth]{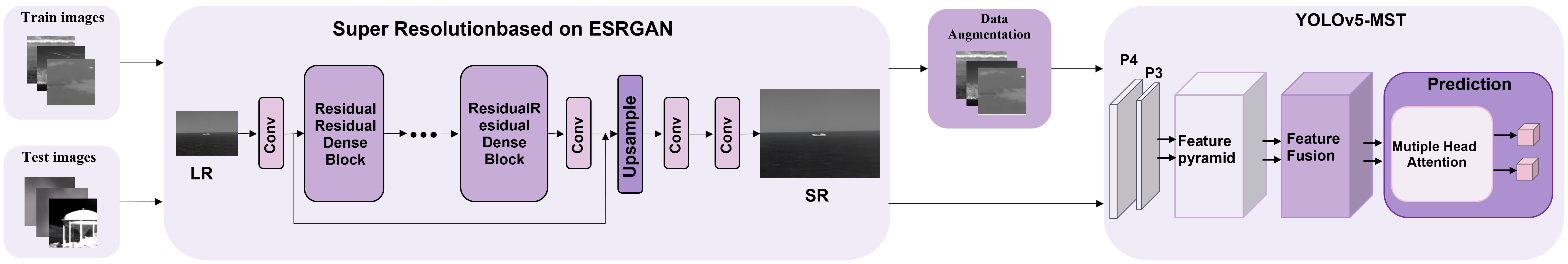}
	\caption{Working diagram overview of the proposed method.}
	\label{fig:Working diagram}
\end{figure*}

\subsection{Segmentation-based methods}   %%2.3%%
%%\label{Semantic Segmentation Methods}
Semantic segmentation is to assign each pixel in an image to a specific category. The objective is to perform pixel-level classification, where pixels belonging to the same category are assigned identical labels. The input features are processed through encoding and decoding operations, which compress and expand the information. This process enables the model to output the target's location and scale information. Given the substantial disparity between the target and its surrounding context, Wang et al. \cite{wang2019miss} proposed MDvsFA, a new infrared small target segmentation framework, which adopts the generative adversarial network (GAN) paradigm to decompose the target segmentation problem into a balance between missed detection and false alarms. Similarly, Wu X et al. \cite{wu2022uiu} proposed a simple and effective "U-Net in U-Net" framework, referred to as UIU-Net, which achieves multi-level and multi-scale representation learning of the object by embedding a smaller U-Net into a larger U-Net backbone network. In addition, Dai et al. \cite{dai2021asymmetric} proposed a typical dataset called SITEMAP and designed an asymmetric context modulation module (ACM) specifically for infrared small target detection to supplement the bottom-up modulation channel. Ren D et al. \cite{ren2021dnanet} proposed a dense nested attention network (DNANet) to directly restore clear images from blurred images through a new connection path topology. In 2024, Xu et al. \cite{yuan2024sctransnet} proposed a single-branch real-time segmentation network SCTransNet, which leverages transformer to align the semantic information of CNNs. This approach retains the fast inference capability of a lightweight, while benefiting from the high accuracy of transformer networks. Additionally, SCTransNet introduces the SIAM module, which effectively aligns features for improved performance. These methods have expanded the methods of semantic segmentation methods in the field of weak infrared small target detection.

\subsection{Attention-based methods}   %%2.4%
%%\label{Attention-based methods}
The Vision Transformer (ViT) effectively performs global image modeling by partitioning the image into fixed-size patches, and input then into the transformer model as a sequence. Zhu X et al. \cite{zhu2020deformable} proposed a relatively new object detection framework, Detection Transformer abbreviated as DETR, which was the first transformer proposed for object detection. The central concept of DETR is to regard the object detection task as a set prediction problem and implement it via the transformer model of the encoder-decoder architecture. This method effectively avoids the post-processing steps relied on in traditional object detection methods, such as non-maximum suppression(NMS) and anchors, thereby simplifying the detection process and improving the overall efficiency and accuracy of the model. In 2021, Liu Z et al. \cite{liu2021swin} proposed Swin Transformer, which is a variant based on Detection Transformer and a layered transformer. Its representation is computed using a shifted window mechanism, which imparts a hierarchical structure to the transformer, enabling it to capture multi-scale features akin to convolutional neural networks (CNNs). This hierarchical framework allows for flexible modeling across different scales while maintaining linear computational complexity with respect to the image size. Zhang et al. \cite{zhang2024global} proposed a global attention network(GANet) based on multi-scale feature fusion, which uses the transformer attention module and the adaptive asymmetric fusion module to detect infrared small targets. Experiments show that this method has high detection accuracy and low false alarm rate on infrared small target datasets.

\section{Methodology}
To balance speed and accuracy in detection tasks, the single-stage YOLOv5 network is adopted as the base framework. Fig. \ref{fig:Working diagram} provides an overview of the proposed method, with the YOLO-MST detection network detailed in Section 3.2. This network comprises three submodules, corresponding to Sections 3.2.2–3.2.4. 
\newline \indent During the data preprocessing stage, we apply data augmentation techniques to the input images to expand the dataset. Additionally, the Real-ESRGAN model \cite{wang2021real} is utilized to greatly enhance the resolution and clarity of the images. Moreover, to reduce missed detection and false positives in the detection results, we replace the SPPF module with the MSFA module in backbone, which improves the feature extraction capability and model operation efficiency under complex backgrounds. During the feature fusion process, we optimize the network model so that the model can focus more on the detection of small infrared targets, and introduce the DyHead module in the prediction head to improve the expression capability of the model's target detection head.

\subsection{Super-Resolution preprocessing and data augmentation}   %%3.1%%
%%\label{Model Overview}
The design and manufacturing constraints of infrared sensors result in lower spatial resolution than those of visible light sensors do, leading to infrared images with reduced resolution and making it challenging to obtain high-quality, high-resolution images. Furthermore, the multiple down-sampling operations in convolutional neural networks exacerbate this issue, as they further degrade image resolution and increase the risk of losing critical feature information, particularly when detecting small targets. To tackle this problem, expanding the pixel resolution of small targets using super-resolution techniques plays a crucial role in improving feature extraction. In 2021, X. Wang et al. \cite{wang2021real} proposed Real-ESRGAN, which can restore images with a relatively high signal-to-noise ratio. Real-ESRGAN not only effectively augments image resolution but also preserves high-frequency details of small targets during the super-resolution process. The version of Real-ESRGAN employed in this study is based on a U-Net discriminator with spectral normalization, with a scaling factor set to 4. The trained model can increase the image resolution by a factor of 4, notably increasing both the clarity and detail fidelity of the image.

\begin{figure*}[htbp]
	\centering
	\includegraphics[width=\textwidth]{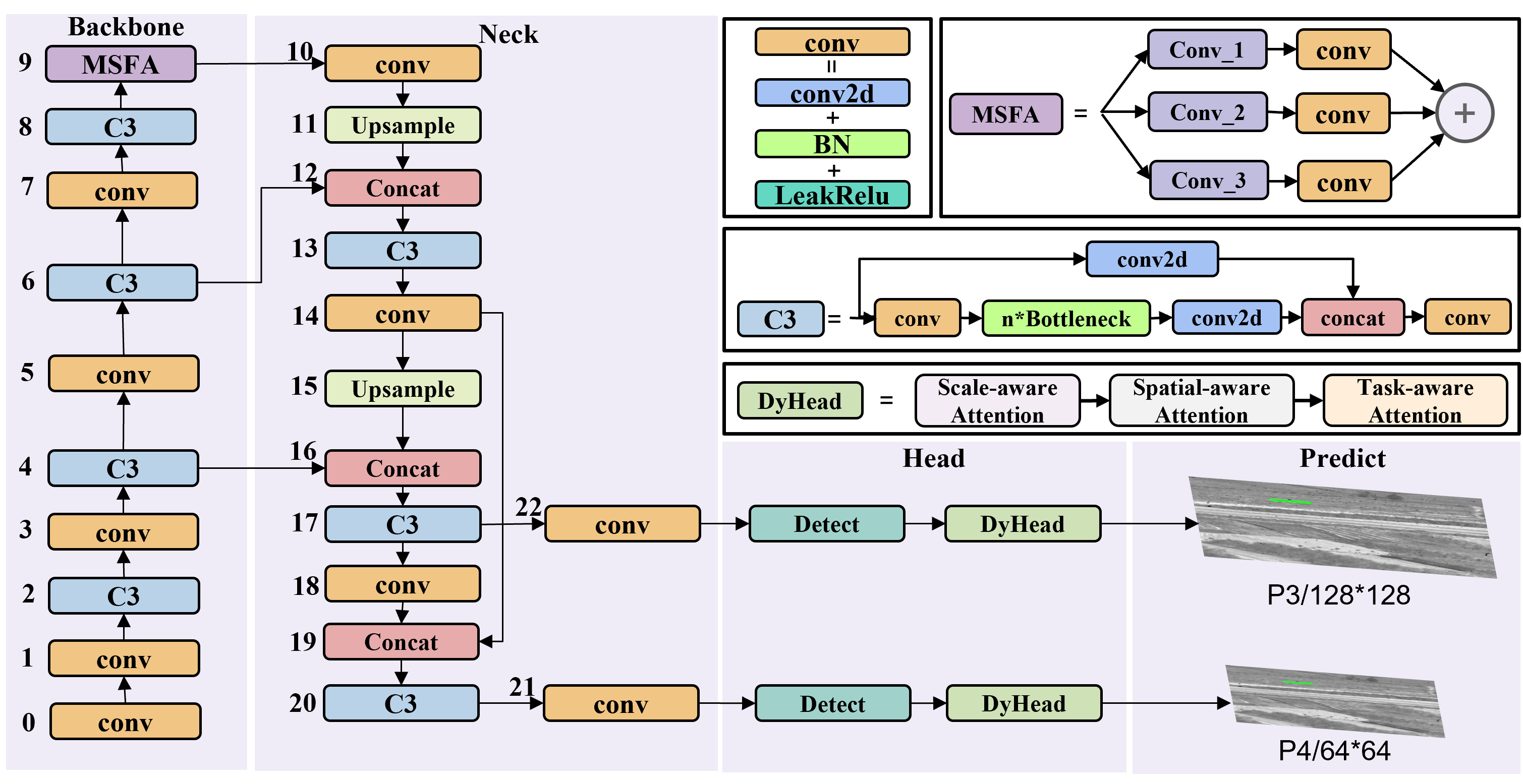}
	\caption{YOLO-MST detection network.} 
	\label{fig:network model}
\end{figure*}

\subsection{YOLO-MST detection network}   %%3.2%%
Considering the speed and accuracy requirements of infrared small target detection tasks, we use the one-stage deep-learning algorithm YOLOv5 network as the basic framework, which mainly include the backbone, neck and prediction head. In reference to the characteristics of infrared small targets, we have improved the three parts of the network and proposed a detection network YOLO-MST, as shown in Fig. \ref{fig:network model} In the backbone part, the MSFA module is adopted in place of the SPPF module, the neck part network structure is optimized, and the DyHead detection head is added to the prediction head \cite{dai2021dynamic}.

\subsubsection{Using MSFA to replace SPPF in backbone}   %%3.2.1%%

The main architecture of the MSFA module refers to the ideas of the CAM module \cite{xiao2022context} and the SPP module \cite{he2015spatial} and is designed on the basis of characteristics of the infrared dataset. The MSFA network combines contextual information enhancement with feature refinement techniques, uses multi-scale dilated convolution to achieve feature fusion, and injects it into the feature pyramid network layer by layer from top to bottom. This method not only effectively improves the feature extraction capability of tiny targets, but also bridges the semantic differences between layers, thereby providing richer contextual information. This design is particularly suitable for infrared small target detection tasks. In addition, the strategy of multi-scale information fusion helps to enhance the model's adaptability to various target sizes and shapes, and improves the recognition capability of small targets at different scales. The calculation formula of the MSFA module is as follows: Formula (1), and the algorithm framework of the MSFA module is shown in Fig. \ref{fig:MSFA} below.

\begin{figure}[htbp]
	\centering
	\includegraphics[width=.5\textwidth]{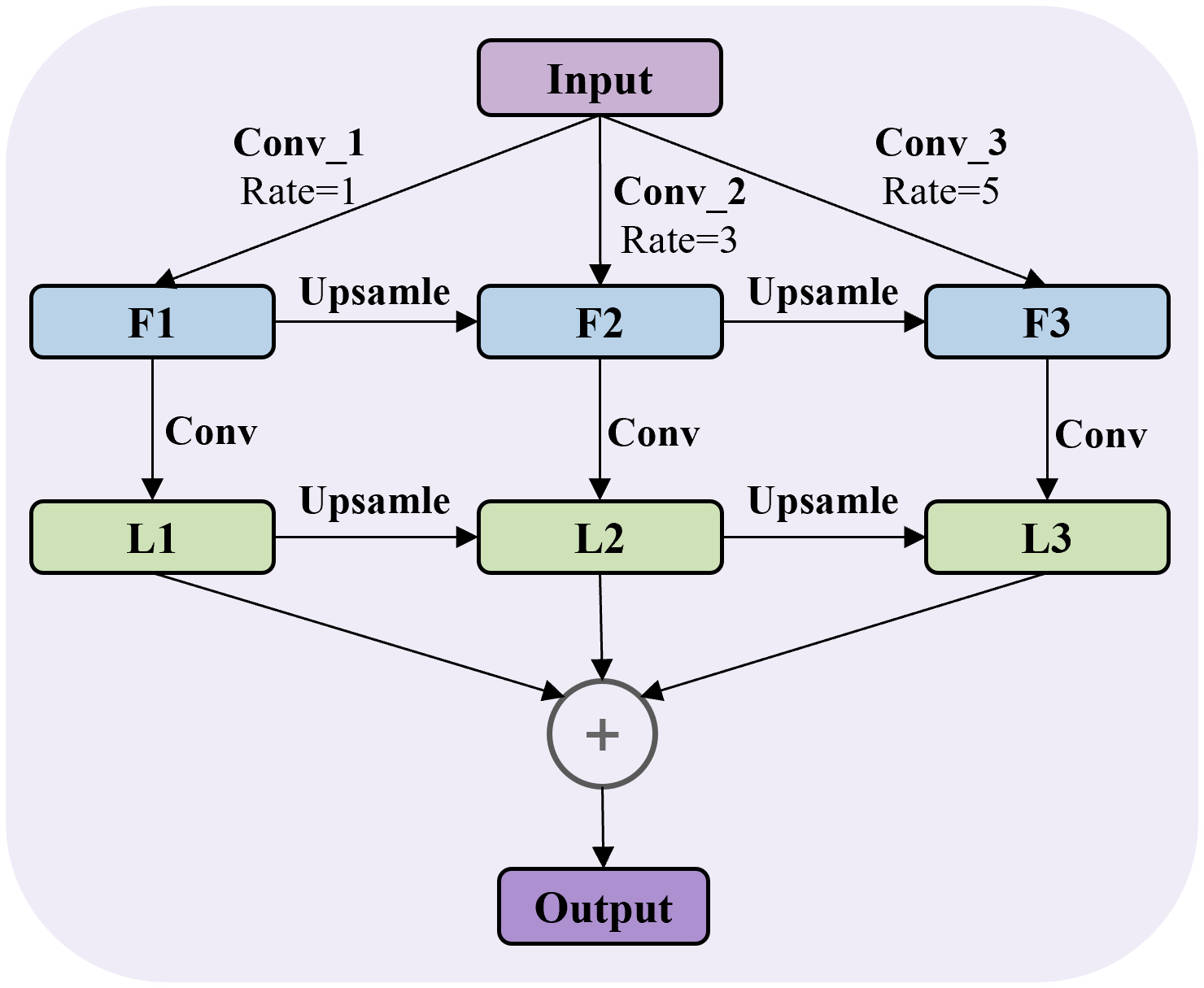}
	\caption{Feature fusion process with MSFA module.} 
	\label{fig:MSFA}
\end{figure}

%\vspace{-20pt}  % 调整这个值来控制间距的大小

\begin{equation}
	%\scriptsize
	\resizebox{0.44\textwidth}{!}{$F_{\text{MSFA}} = \text{Conv}(Conv_{\_}1(F)) + \text{Conv}(Conv_{\_}2(F)) + \text{Conv}(Conv_{\_}3(F))$}
\end{equation}

$Conv(\cdot)$ represents a convolution operation with a $1 \times 1$ kernel. $Con_{\_}1(\cdot)$, $Con_{\_}2(\cdot)$, and $Con_{\_}3(\cdot)$ represent dilated convolutions with dilation rates of 1, 3, and 5, respectively. $ F $ and $F_{\mathrm{MSFA}}$ are the input and output features of the MSFA module, respectively.

\subsubsection{Optimization of the neck structure}   %%3.2.2%%
\begin{figure}[htbp]
	\centering
	%\raggedright
	\subfloat{
		\includegraphics[width=3.5cm]{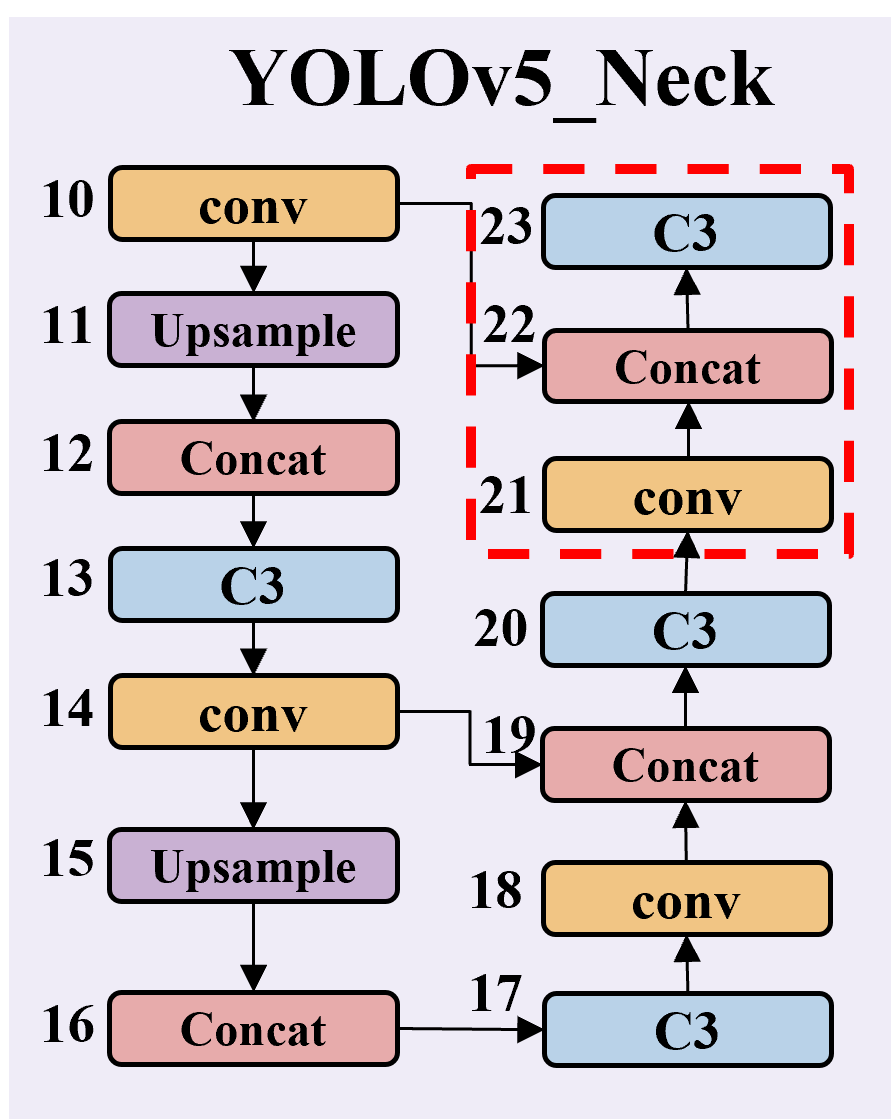}}
	\subfloat{
		\includegraphics[width=4.85cm]{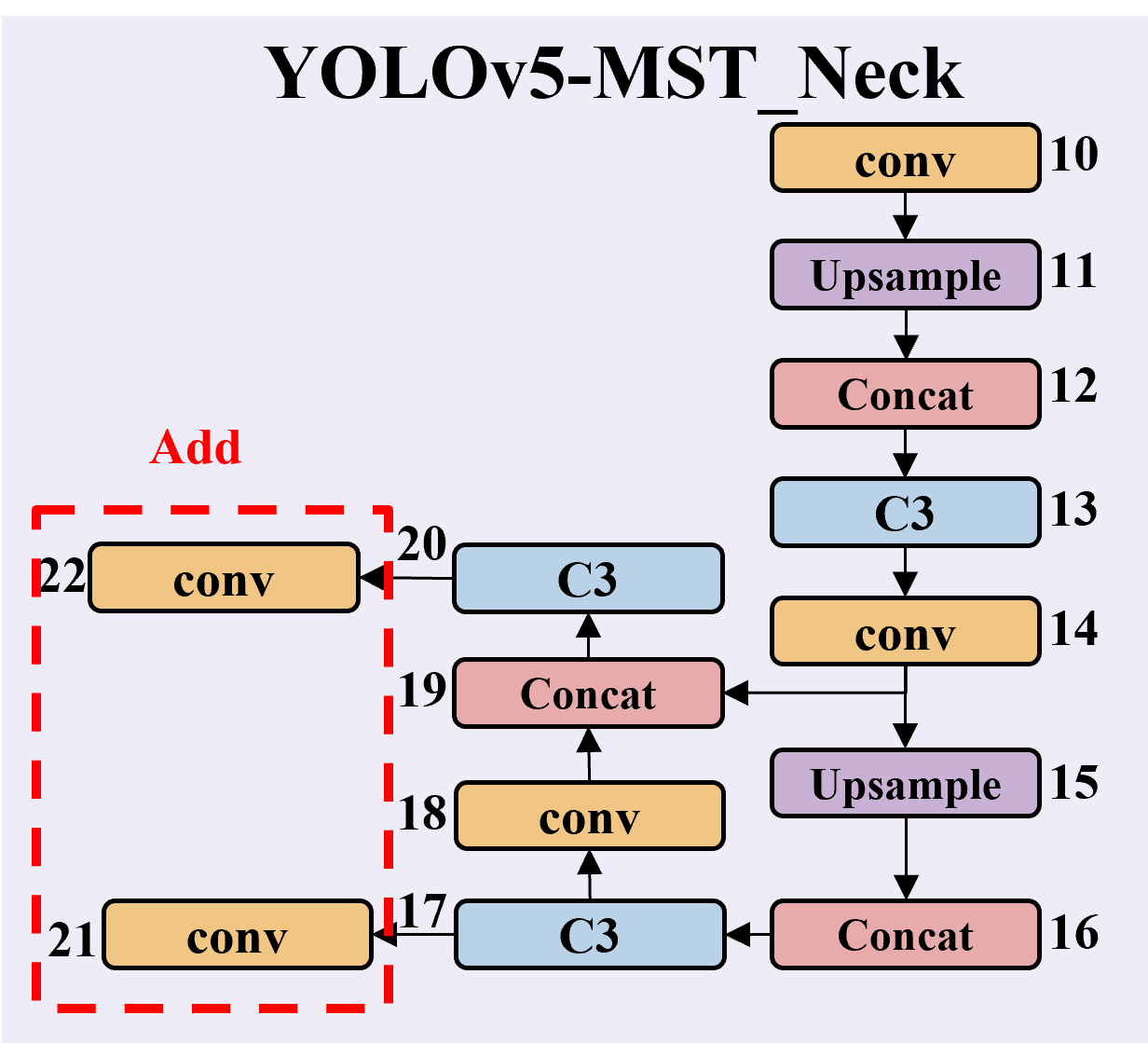}}
	\caption{Comparison of the neck network structure before and after optimization: the left side represents the original structure, whereas the right side illustrates the improved version.}
	\label{fig:neck}
\end{figure}

The output component for large target detection is removed from the neck, and an additional convolutional layer is inserted between the two outputs. This modification optimizes the neck structure of the model, allowing it to better focus on small target detection.
\newline \indent The neck component of the YOLOv5 model produces three outputs for small, medium, and large targets. Nonetheless, the large target output can interfere with the small target output \cite{wang2021multi}. To address this issue, we optimize the network structure of the neck component. The original structure of the neck component is illustrated on the right side of Fig. \ref{fig:neck} The specific optimization includes removing the original 21-layer convolution layer, 22-layer concatenation layer, and 23-layer C3 module, as well as the subsequent part of the 17th and 20th layer outputs of the network, and adding a convolution layer for further feature extraction and processing. The optimized network structure of the neck part is shown on the left side of Fig. \ref{fig:neck}, which ensures more effective integration and output of features.

%\vspace{-10pt}

\subsubsection{Adding DyHead to prediction heads}   %%3.2.3%%
\begin{figure}[htbp]
	\centering
	\includegraphics[width=.48\textwidth]{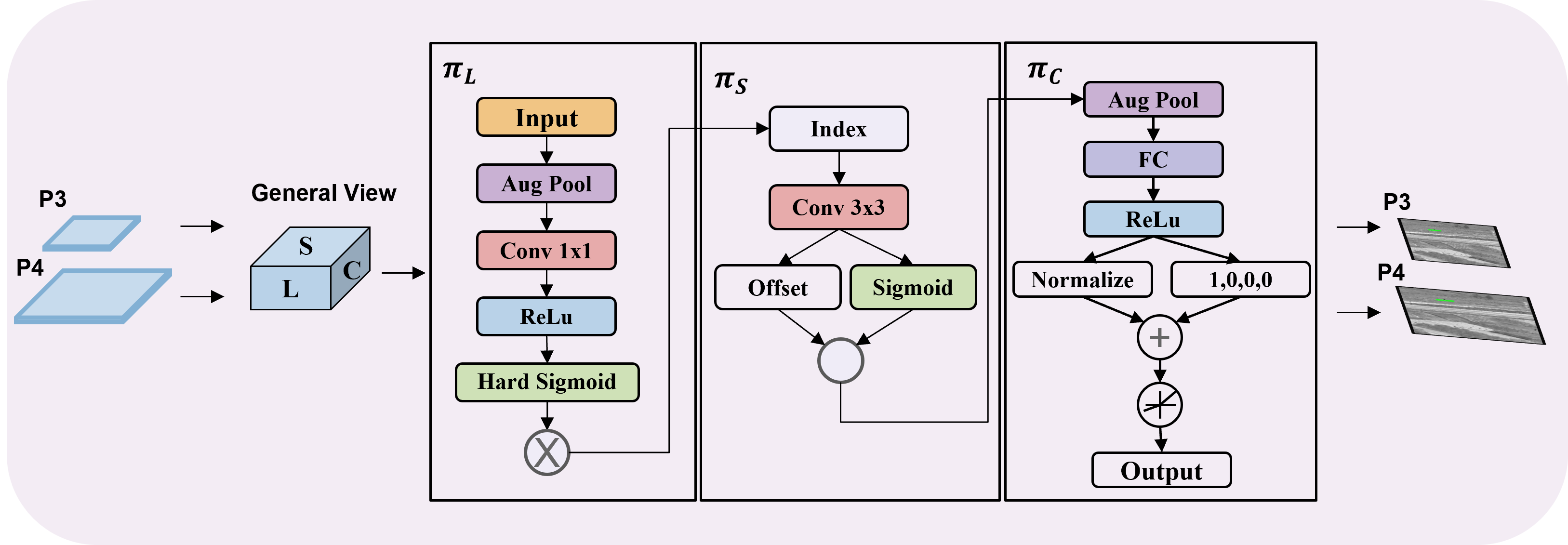}
	\caption{DyHead structure diagram.} 
	\label{fig:dyhead}
\end{figure}

%\label{}
At the end of the target detection network, the prediction head regresses and predicts the target using prior boxes. However, this approach often neglects the feature relationships between small targets and their surrounding context, limiting the network's overall understanding of the image. Additionally, the lack of multi-scale features hinders the network’s ability to capture important details, particularly when handling targets of varying sizes, resulting in reduced detection performance. This limitation affects YOLOv5's robustness in complex infrared environments. To address these challenges and improve the network's contextual awareness—especially in capturing multi-scale information—we introduce DyHead, a multi-scale prediction head with an attention mechanism. The schematic diagram of DyHead is shown in Fig. \ref{fig:dyhead}.

%\vspace{-10pt}

Given a feature tensor $F\in R^{L*S*C}$,the attention function is converted into three consecutive attentions, each focusing on only one perspective:
\begin{equation}
	W(F)=\pi_{C}(\pi_{S}(\pi_{L}(F)\cdot F)\cdot F)\cdot F 
\end{equation}
Where $\pi_{L} (\cdot) $, $\pi_{S} (\cdot)$ and $\pi_{C} (\cdot)$ are three different attention functions applied to dimensions $L, S$ and $C$ respectively. $\pi_{L} (\cdot)$ is the scale-aware attention function, $\pi_{S} (\cdot)$ is the space-aware attention function, and $\pi_{C} (\cdot)$ is the task-aware attention function. The function expressions are as follows (2)-(4)
\begin{equation}
	\pi_{L}(F)\cdot F=\sigma(f(\frac{1}{SC}\sum_{S}^{C}F))\cdot F 
\end{equation}
Here, $f(\cdot)$ is a linear function approximated by a 1×1 convolutional layer, %$\sigma(x)=max(0,min(1,\frac{x+1}{2})) $ is the hard-sigmoid function.
$\sigma(x) = \max\left( 0, \min\left( 1, \frac{x+1}{2} \right) \right) $ is the hard-sigmoid function.
\begin{equation}
	\pi_{S}(F)\cdot F=\dfrac{1}{L}\sum_{l=1}^{L}\sum_{k=1}^{K}W_{1,K}\cdot F(l;P_{k}+\Delta P_{k};C)\cdot\Delta m_{k}
\end{equation}
Where K is the number of coefficient sampling locations, $P_{k}+\Delta P_{k}$ is the location shifted by the self-learned spatial offset $\Delta P_{k}$ to focus on the discriminative region, and $\Delta m_{k}$ is the self-learned importance scalar at location $P_{k}$. Both are learned from the input features at the median level of F.
\begin{equation}
	\pi_{C}(F)\cdot F=max(\alpha^{1}(F)\cdot F_{c}+\beta^{1}(F),\alpha^{2}(F)\cdot F_{c}+\beta^{2}(F))
\end{equation}
Where $F_{c}$ is the feature slice of the cth channel, [$\alpha^{1},\alpha^{2},\beta^{1},\beta^{2}] \\= \theta(\cdot$) is a hyperfunction that learns to control the activation threshold. Implementation of $\theta(\cdot)$: It first performs global average pooling is performed on the $L\times S$ dimension to reduce the dimensionality, then uses two fully connected layers and a normalization layer, and finally applies a shifted sigmoid function to normalize the output to [-1, 1].
Finally, since the above three attention mechanisms are applied sequentially, we can nest Equation (2) multiple times to effectively stack multiple $\pi_{L}(\cdot)$, $\pi_{S}(\cdot)$, and $\pi_{C}(\cdot)$ blocks together.

\section{Experimental results}

\subsection{Dataset preparation}   %%4.1%%
%%\label{}
To evaluate the effectiveness of this method in detecting infrared small targets across different scales, two publicly available infrared small target datasets were selected for this study. These two sets of datasets are the Infrared Image Sequence Dataset (IRIS) for Aircraft Small Target Detection and Tracking \cite{hui2020dataset} and the Single Frame Infrared Small Target Dataset (SIRST) \cite{dai2021asymmetric}.
\newline \indent The IRIS dataset encompasses various tracking challenges, including scale changes, deformation, fast motion, motion blur, background clutter, and low resolution. Each image has a resolution of 256×256 pixels. However, the original dataset, annotated using electrical labeling methods, is unsuitable for deep-learning training. To resolve this, 1,697 images were manually annotated with high-quality labels using LABELIMG software, enabling their use in deep-learning models. The dataset is divided into training, validation, and test sets at a 6:2:2 ratio, containing a total of 1,738 targets, with each target occupying approximately 0.12\% of the image area.
\newline \indent In the SIRST dataset, 90\% of the images have only one target, and 10\% of the images have multiple targets; 55\% of the target area accounts for less than 0.02\% (that is, in a 300*300 image, the target pixel is 3*3). SIRST uses the bounding box annotation method, including a total of 427 infrared images and 480 targets. The dataset is randomly divided into training: validation: testing = 3:1:1.

\begin{figure*}[!htb]
	\centering
	\includegraphics[width=\textwidth]{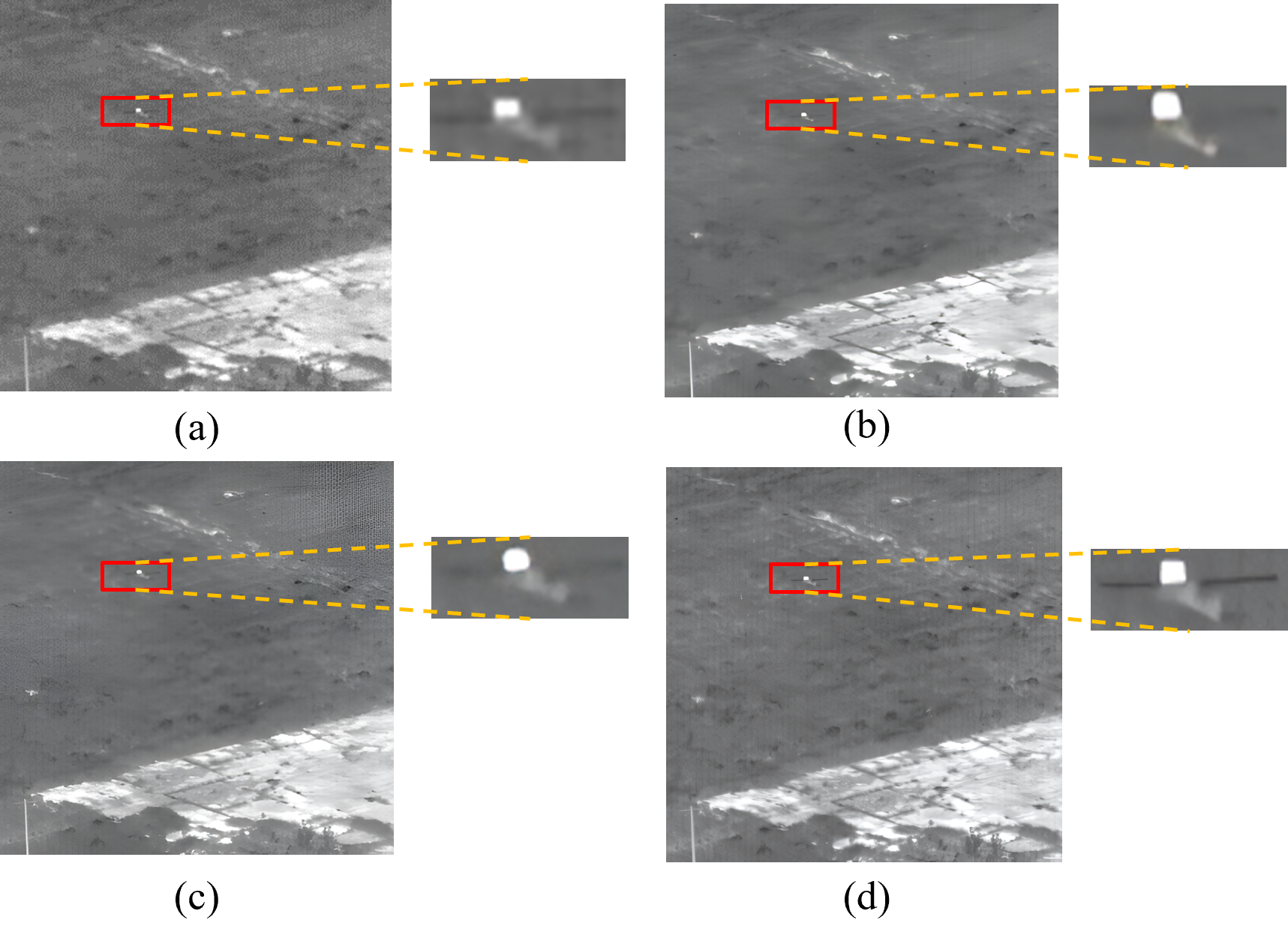}
	\caption{Examples of different super-resolution preprocessing methods: a) original image, b) IERN \cite{chen2021lightweight}, c) SRResNet \cite{ledig2017photo}, d) ESRGAN \cite{wang2021real}.} 
	\label{fig:SRR}
\end{figure*}

\subsection{Experimental details}   %%4.2%%
%%\label{}
All the experiments were conducted on an Intel(R) Xeon(R) Gold 5218 CPU, Linux operating system, and 24.0 GB of RAM. All the deep-learning models used were based on the Torch framework. The GPU graphics card used to accelerate training was (NVIDIA) GeForce GTX 3090 Ti, and the Cuda version was 12.1.
\newline \indent The ESRGAN network was applied to achieve super-resolution for all infrared image data, and the infrared images were employed to retrain the ESRGAN and generate new super-resolution images for subsequent model training and testing. The experimental results of the three super-resolution methods are shown in Fig. \ref{fig:SRR}.
The Adam optimizer with a smoothing constant of 0.9 is adopted to optimize the MSE loss function. The learning rate of the training is $10^{-3}$, the iterative update size of the learning rate is $10^{-2}$, and the batch size is 48. Referring to the XML annotation of the original image, we use LABELIMG pairs for manual annotation.
\newline \indent The test experiments were conducted on the test sets of IRIS and SIRST. The input image size is $1280 \times 1280$, the batch size is set to 48, the IoU threshold of NMS is 0.45, the confidence threshold of the predicted box is 0.15, and other parameters are set to the default values of YOLOv5.

\subsection{Comparison with other advanced methods}   %%4.3%%
%%\label{}
The proposed method is tested on two public datasets. YOLO-MST achieves 99.5\% and 96.4\% mAP@0.5 on the IRIS dataset and 70.5\% and 47.5\% mAP@0.95 on the SIRST dataset. Tables \ref{tab:iris} and \ref{tab:sirst} show the comparison of the test results on the two datasets, Table \ref{tab:yolo} lists the comparison of various target detection methods in terms of parameters, model size, computational effort (FLOPs), detection accuracy (mAP@0.5 and mAP@0.5:0.95). These methods encompass both general deep-learning-based target detection algorithms, such as the YOLO series and CNN-based approaches, along with their respective variations incorporating attention mechanisms. Additionally, special methods developed in recent years for infrared small target detection, such as SRResNet + YOLOSR-IST-s, are also included.
\newline \indent The experimental results show that in both experimental datasets, YOLO-MST has good infrared target detection performance in both detection paradigm-based methods and segmentation paradigm-based methods. Specifically, on the IRIS dataset, YOLO-MST achieves 99.5\% in mAP, among the various performance indicators, it achieves the highest score in terms of comprehensive detection capability, surpassing other methods, while maintaining an acceptable FPS range. For the SIRST dataset, YOLO-MST outperforms the other methods in both precision and recall. This finding demonstrates that YOLO-MST effectively reduces false alarms, and significantly minimizes the omission of true targets, which is a common issue with the comparison methods. Furthermore, YOLO-MST delivers excellent detection performance for small targets across different scales in both datasets, underscoring its robust performance and adaptability.

\begin{table}[!htb] 
	\small
	\centering      
	\caption{Comparison of test results on the IRIS dataset(target size$\geq$0.12\%, 200 epochs)}
	\label{tab:iris} 
	\begin{tabular}{@{}ccc@{}} 
		\toprule
		Detection method & mAP@0.5/\% & FPS \\  
		\midrule
		Faster R-CNN \cite{lee2017me} & 59.2 & 12.7 \\ %\hline
		RetinaNet \cite{lin2017focal} & 72.1 & 49.4 \\ %\hline
		YOLOv3-tiny \cite{farhadi2018yolov3} & 78.3 & 76.1 \\ %\hline
		RFBnet \cite{liu2018receptive} & 80.1 &  65.4 \\ %\hline
		YOLOv5s & 88.3 &   71.5 \\ %\hline
		ISTDet+ Image Filtering Module \cite{ju2021istdet} & 94.3 &   77.2 \\ %\hline
		YOLO-SASE + SRGAN \cite{zhou2022yolo}  & 96.8 &  51.9 \\ %\hline
		YOLOv6-S \cite{li2022yolov6} & 92.4 &   44.7 \\ %\hline
		YOLOv7-W6 \cite{wang2023yolov7} &  91.8 &  53.9 \\ %\hline
		YOLOv8s & 94.3 & 46.7 \\ %\hline
		YOLOv9-S & 93.5 & 44.1 \\ %\hline
		YOLOv10s \cite{wang2024yolov10} & 95.5 & 47.9 \\ %\hline
		SRResNet + YOLOSR-IST \cite{li2023yolosr} & 99.2 & 45.2 \\ %\hline
		\textbf{ESRGAN+ YOLO-MST(Ours)} & \textbf{99.5} & \textbf{52.7}\\ 
		\bottomrule
	\end{tabular}
\end{table}
%\vspace{1em} 
%\vspace{-10pt}

\begin{table}[!htb] 
	\small
	\centering     
	\caption{Comparison of test results on the SIRST dataset(target size$\leq$0.02\%, 250 epochs)}
	\label{tab:sirst}
	\begin{tabular}{@{}ccc@{}} 
		\toprule
		Detection method &Precision/\% & Recall/\% \\ 
		\midrule
		Faster R-CNN \cite{lee2017me} & 53.2 & 46.7 \\ %\hline
		RetinaNet \cite{lin2017focal} & 61.5 & 59.9 \\ %\hline
		YOLOv3-tiny \cite{farhadi2018yolov3} & 65.4 & 60.7 \\ %\hline
		RFBnet \cite{liu2018receptive} & 58.6 &  57.5 \\ %\hline
		YOLOv5s & 73.3 &   66.3 \\ %\hline
		MdvsFA \cite{wang2019miss} & 84.6 &   84.5 \\ %\hline
		ACM \cite{dai2021asymmetric}  & 88.3 &  89.9 \\ %\hline
		ALCNet \cite{dai2021attentional}  & 86.6 &  87.3 \\ %\hline
		YOLOv6-S \cite{li2022yolov6} & 64.5 &   61.1 \\ %\hline
		YOLOv7-W6 \cite{wang2023yolov7}  &  65.1 &  62.6 \\ %\hline
		YOLOv8s & 71.6 & 71.1 \\ %\hline
		YOLOv9-S & 68.7 & 72.3 \\ %\hline
		YOLOv10s \cite{wang2024yolov10} & 63.1 & 73.5 \\ %\hline
		SRResNet + YOLOSR-IST \cite{li2023yolosr} & 93.7 & 91.0 \\ %\hline
		\textbf{ESRGAN+ YOLO-MST(Ours)} & \textbf{95.8} & \textbf{91.8}\\ 
		\bottomrule	
	\end{tabular}
\end{table}
%\vspace{1em} 
%\vspace{-10pt}

%\iffalse
\begin{table*}[!htb]
	\small
	\centering     
	\caption{Comparison of different YOLO model indicators on the SIRST dataset}
	\label{tab:yolo}
	\resizebox{\textwidth}{!}{% 
		\begin{tabular}{@{}cccccc@{}}
			\toprule
			Detection method & Params(M) & Model size(MB) & Flops(G) & mAP@0.5/\% & mAP@0.5:0.95/\% \\  
			\midrule
			YOLOv3-tiny \cite{farhadi2018yolov3} & 12.1 & 24.4 & 18.9 & 61.4 & 23.9 \\ %\hline
			YOLOv5s & 9.1 & 18.5 & 23.3 & 69.2 & 29.7 \\ %\hline
			YOLOv6-S \cite{li2022yolov6} & 16.3 & 31.3 & 44.0 & 68.5 & 28.3 \\ %\hline
			YOLOv7-W6 \cite{wang2023yolov7} & 141.9 & 135.3 & 105.1 & 65.1 & 26.1 \\ %\hline
			YOLOv8s & 11.1 & 14.4 & 28.4 & 73.8 & 30.7 \\ %\hline
			YOLOv9-S & 12.8 & 26.3 & 59.3 & 65.4 & 26.3 \\ %\hline
			YOLOv10s \cite{wang2024yolov10} & 8.0 & 16.6 & 24.5 & 63.4 & 24.8 \\ %\hline
			SRResNet + YOLOSR-IST-s \cite{li2023yolosr} & 6.4 & 13.2 & 12.7 & 94.6 & 46.3 \\ %\hline
			\textbf{ESRGAN+ YOLO-MST(Ours)} & \textbf{12.7} & \textbf{25.9} & \textbf{20.9} & \textbf{96.4} & \textbf{47.5} \\ 
			\bottomrule	
		\end{tabular}
	}
\end{table*}
%\fi

\subsection{Ablation experiment}   %%4.4%%
Table \ref{tab:ablation} shows the results of the YOLO-MST ablation experiment (this experiment is based on the SIRST dataset).
\newline \indent The results in Table \ref{tab:ablation} show that ESRGAN+YOLO-MST improves the original YOLOv5-s mAP$@$0.5 by $27.2\%$ and mAP$@$0.5:0.95 by $17.8\%$. Among them, super-resolution ESRGAN is the key element of YOLOSR-IST, which improves the model's mAP$@$0.5 by $16.9\%$ and mAP$@$0.5:0.95 by $11.9\%$. In contrast, the detection performance of the YOLO-v5-s model using only super-resolution is not ideal. Firstly, this limitation stems from the insufficient scale of the training dataset. Secondly, YOLOv5's multi-scale feature fusion fails to fully capture the critical spatial features of the target and does not effectively exploit the contextual relationships between different local features during processing. Moreover, there is a lack of effective strategies to better focus on the region of interest and address the issue of feature loss caused by the oversampling of small targets. To address these problems, our solutions include using the innovative MSFA module in the backbone part to capture richer multi-scale feature information of the image. Optimize the network structure of the neck. Adding DyHead to the prediction head enables the model to be more focused on the shape of the object and enhances feature extraction.
\newline \indent The results of the ablation experiments demonstrate that the ESRGAN-based super-resolution preprocessing, data augmentation, neck optimization, MSFA, and DyHead modules work synergistically to enhance detection performance. Each additional improvement module contributes to a progressive increase in the model's mAP, highlighting the effectiveness of these components in optimizing the detection accuracy. These findings further validate the robustness and scientific foundation of the detection approach proposed in this study.
%\balance

\begin{table*}[!htb]
	\small
	\centering   
	\caption{Ablation test results of YOLO-MST}
	\label{tab:ablation}
	\resizebox{\textwidth}{!}{% 
		\begin{tabular}{@{}cccccccc@{}}
			\toprule
			YOLOv5s & Super-resolution & Data Augmentation & MSFA & Neck Optimize & DyHead & mAP@0.5/\% & mAP@0.5:0.95/\% \\  
			\midrule
			\checkmark &  &  &  &  &  & 69.3 & 29.7 \\ %\hline
			\checkmark & \checkmark &  &  &  &  & 86.1(+16.9) & 41.6(+11.9) \\ %\hline
			\checkmark & \checkmark & \checkmark &  &  &  & 88.4(+2.3) & 41.8(+0.2) \\ %\hline
			\checkmark & \checkmark & \checkmark & \checkmark &  &  & 92.1(+3.7) & 44.2(+2.4) \\ %\hline
			\checkmark & \checkmark & \checkmark & \checkmark & \checkmark &  & 93.9(+1.8) & 45.5(+1.3) \\ %\hline
			\checkmark & \checkmark & \checkmark & \checkmark & \checkmark & \checkmark & 96.4(+2.5) & 47.5(+2.0) \\ 
			\bottomrule
		\end{tabular}
	}
\end{table*}
%\FloatBarrier  % 强制所有浮动体排布完
%\clearpage
%\vfill

\subsection{Visualization of test results}   %%4.5%%
\begin{figure*}[!htb]
	\centering
	\includegraphics[width=\textwidth]{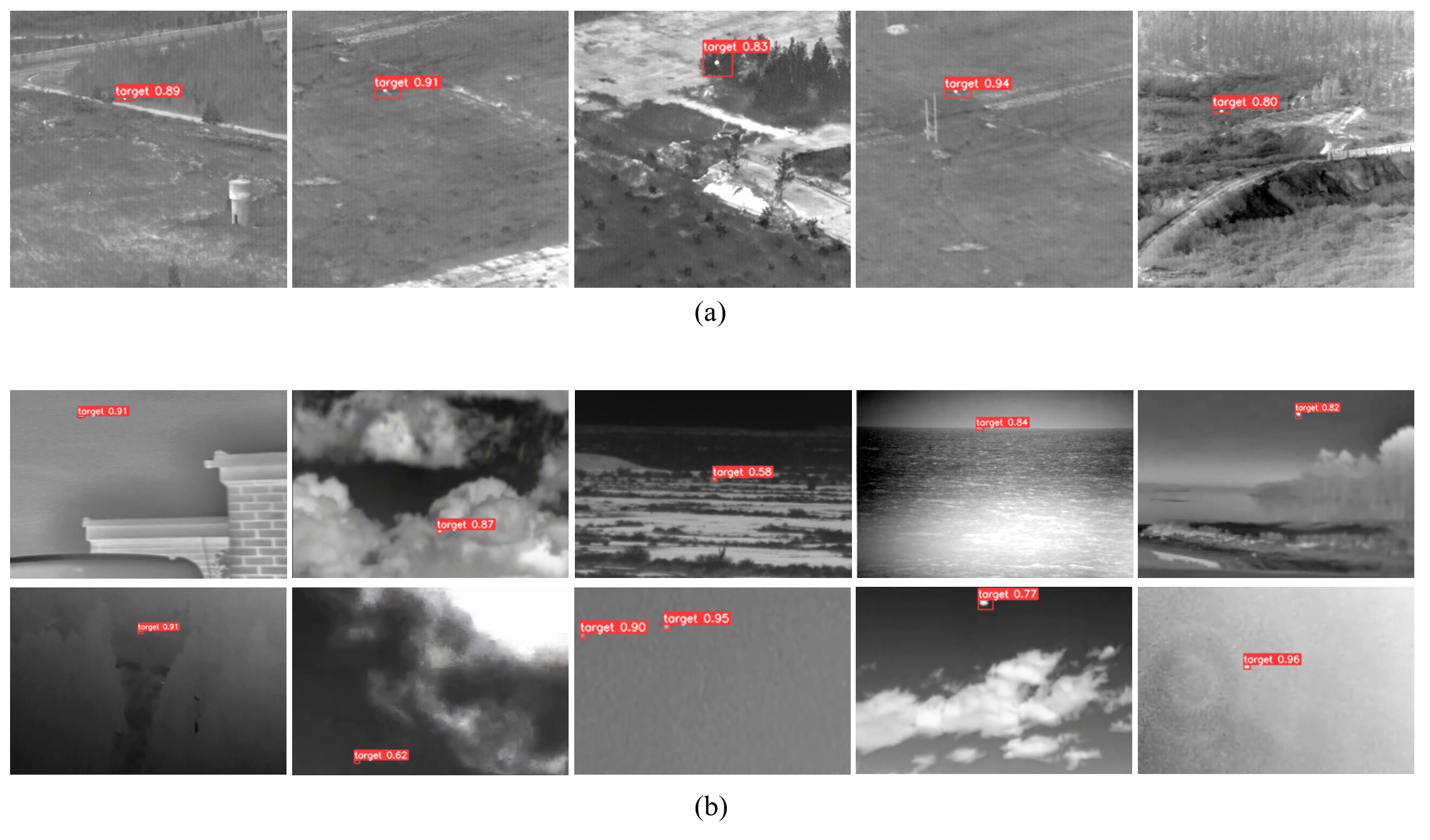}
	\caption{Examples of visualization results on test set detection, covering complex imaging backgrounds: a) IRIS dataset, and b) SIRST dataset.} 
	\label{fig:Background}
\end{figure*}

Fig. \ref{fig:Background} shows some representative image visualization results of the test set detection in the IRIS and SIRST datasets. These images cover complex imaging backgrounds such as land, city, sky, ocean, and the transition area between sea, land, and sky. The experimental results show that this method can accurately identify small infrared targets in complex backgrounds, and the confidence of the predicted box is high.

%\FloatBarrier  % 强制所有浮动体排布完
%\nopagebreak

\begin{figure*}[!htb]
	\centering
	\subfloat{
		\includegraphics[width=0.9\textwidth,height=0.4\textwidth]{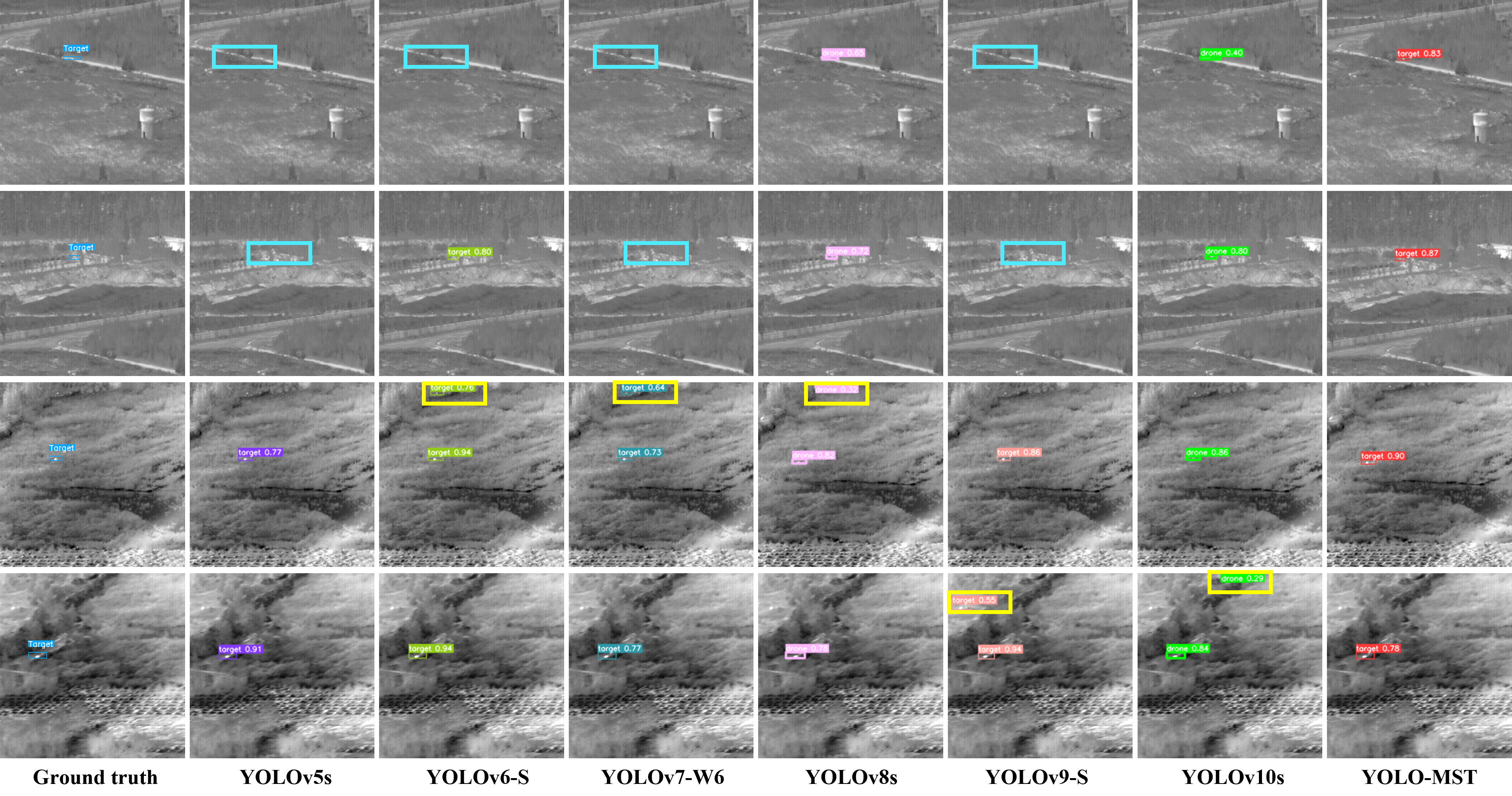}}
		\caption{Comparison results on the IRIS dataset. False positives and false negatives are marked with blue and yellow boxes, respectively.} 
		\label{fig:iris}
		%\\
	\subfloat{
		\includegraphics[width=0.9\textwidth,height=0.4\textwidth]{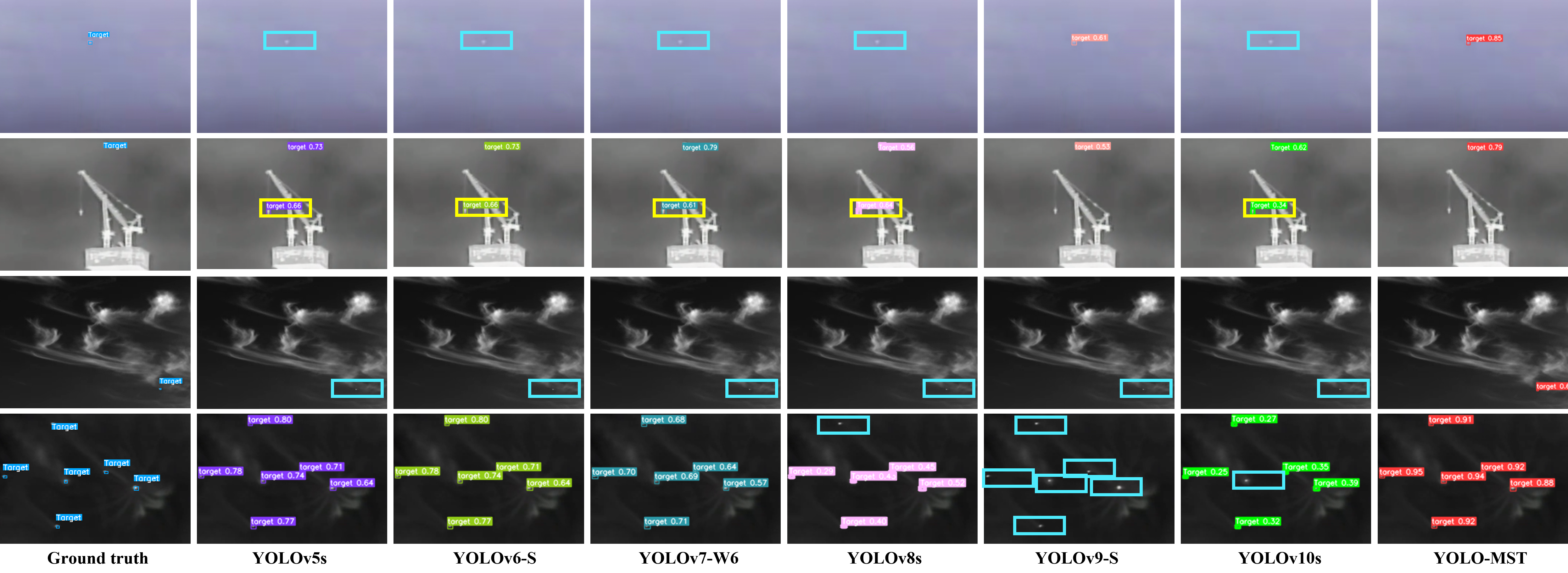}}
		\caption{Comparison results on the SIRST dataset. False positives and false negatives are marked with blue and yellow boxes, respectively.} 
		\label{fig:sirst}
\end{figure*}

\begin{figure*}[!htb]
	\centering
	\subfloat{
		\includegraphics[width=8cm]{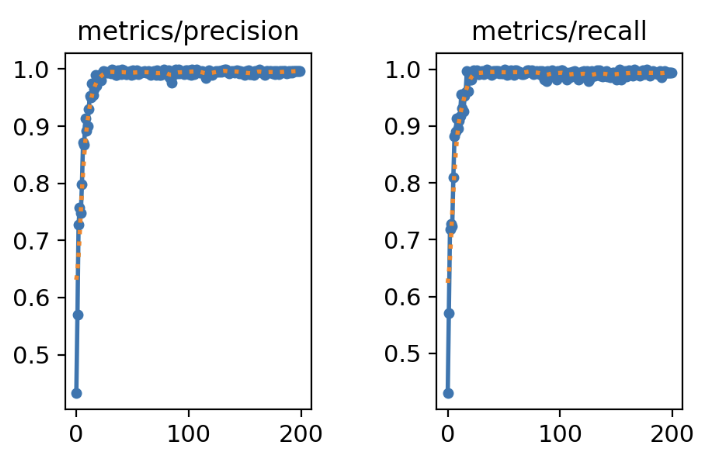}}
	\subfloat{
		\includegraphics[width=8cm]{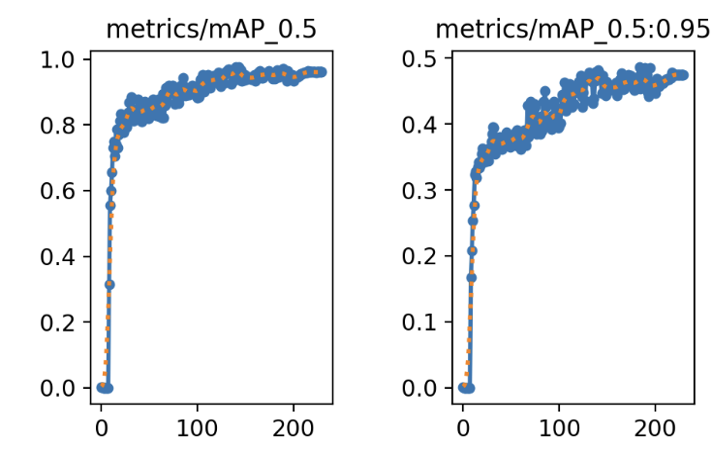}}
	\caption{Training metric results in two datasets. (a) Precision in the IRST dataset, (b) Recall in the IRST dataset, (c) mAP@0.5 in the SIRST dataset, (d) mAP@0.5:0.95 in the SIRST dataset.}
	\label{fig:result}
\end{figure*}

Fig. \ref{fig:iris} and Fig. \ref{fig:sirst} show several representative images from the test dataset as examples to visualize the detection results. The experimental results indicate that the YOLO series algorithms exhibit varying degrees of missed detections and false positives, with some models performing better than others. Specifically, while YOLOv5, YOLOv6, and other variants achieve reasonable accuracy, they still face challenges with false positives and missed detections, especially in complex scenes. In contrast, YOLO-MST successfully avoids both missed detections and false positives, demonstrating competitive detection results in terms of accuracy, recall, and mAP@0.5.
\newline \indent Fig. \ref{fig:result} displays the training results of the two datasets, involving the curves of the two evaluation indicators on the validation set. For the IRIS dataset, YOLOSR-IST can quickly reach a convergence state within the first 30 epochs, and the curve fitting effect is also very good. For the SIRST dataset, the mAP curve of YOLOSR-IST has severe jitter for a period of time after the warm-up process, but through continuous parameter optimization using the cosine annealing algorithm, allowing the algorithm to converge to a stable state within 300 epochs. This convergence not only enhances performance but also improves the robustness of the model in various scenarios.
%\balance
%\vfill
\nopagebreak

%\vspace{-40pt}
%\begin{comment}
%\end{comment}
%\clearpage  % 强制所有浮动对象在这个位置输出  

\section{Summary and conclusions}   %%5%%
This paper proposes a deep-learning infrared small target detection method that integrates image super-resolution technology and multi-scale observation. Initially, the input infrared image undergoes super-resolution preprocessing and multiple data augmentation techniques. Subsequently, we design a deep-learning network, YOLO-MST, on the basis of YOLO model. The MSFA module we developed to replace the SPPF module in the backbone, enhancing feature extraction capabilities and improving model efficiency in complex backgrounds. In the neck part, we eliminate the large target detection output and introduce an additional Conv layer to optimize the structure, enabling the model to prioritize small target detection. Moreover, we incorporate a DyHead module into the head section, which improves the model's target detection capabilities while maintaining computational efficiency.
\newline \indent The experimental results show that by improving the quality of remote sensing images and designing a deep-learning model specifically, the mAP$@$0.5 matching rate of this method on the IRIS dataset reaches 99.5\%, and the precision and recall rates on the SIRST dataset are 95.8\% and 91.8\%, respectively. The comparison results across two public datasets demonstrate that the proposed method outperforms existing approaches in all evaluated metrics. Furthermore, it effectively addresses the issues of missed detection and false positives, which are prevalent in many current detection methods. In future work, we will focus on optimizing the overall network architecture and improving the efficiency of super-resolution image processing to further enhance performance.

\bibliographystyle{cas-model2-names}
\bibliography{cas-refs.bib}

\end{document}